\documentclass[sigconf, natbib=false]{acmart}
\usepackage{enumitem}
\usepackage{booktabs}
\usepackage{graphicx}
\AtBeginDocument{%
  }

\setcopyright{acmlicensed}
\setlist[itemize]{topsep=0pt}
\copyrightyear{2026}
\acmYear{2026}
\acmDOI{XXXXXXX.XXXXXXX}
\acmConference[RecSys '26]{RecSys '26}{Sept 28–Oct 2 2026}{Minneapolis, Minnesota, USA}
\acmISBN{978-1-4503-XXXX-X/2018/06}

\RequirePackage[
  datamodel=acmdatamodel,
  style=acmnumeric,
  ]{biblatex}

\begin{document}

\title{Lightweight Ranking Heads: Accelerating Multi-Task Experimentation in Production Recommender Systems}

\author{Sanjay Surendranath Girija}
\orcid{0009-0004-1759-5183}
\affiliation{%
  \institution{Google LLC}
  \city{Mountain View}
  \state{CA}
  \country{USA}
}
\email{sanjaysg@google.com}

\author{Aniruddh Nath}
\orcid{0009-0006-2815-3035}
\affiliation{%
  \institution{Google LLC}
  \city{Mountain View}
  \state{CA}
  \country{USA}
}
\email{aniruddhnath@google.com}

\author{Li Wei}
\orcid{0009-0008-9321-3983}
\affiliation{%
  \institution{Google LLC}
  \city{Mountain View}
  \state{CA}
  \country{USA}
}
\email{liwei@google.com}

\author{Yanhao Jiang}
\orcid{0009-0004-4150-2020}
\affiliation{%
  \institution{Google LLC}
  \city{Mountain View}
  \state{CA}
  \country{USA}
}
\email{yanhaojiang@google.com}

\author{Shawn Andrews}
\orcid{0009-0004-1129-8196}
\affiliation{%
  \institution{Google LLC}
  \city{Mountain View}
  \state{CA}
  \country{USA}
}
\email{shawnandrews@google.com}

\author{Lukasz Heldt}
\orcid{0009-0003-0593-7345}
\affiliation{%
  \institution{Google LLC}
  \city{Mountain View}
  \state{CA}
  \country{USA}
}
\email{heldt@google.com}

\author{Yi Wu}
\orcid{0009-0005-9363-2086}
\affiliation{%
  \institution{Google LLC}
  \city{Mountain View}
  \state{CA}
  \country{USA}
}
\email{wuyish@google.com}

\author{Aditya Mahajan}
\orcid{0009-0009-3916-5578}
\affiliation{%
  \institution{Google LLC}
  \city{Mountain View}
  \state{CA}
  \country{USA}
}
\email{admahajan@google.com}

\author{Mohit Sharma}
\orcid{0009-0007-4652-0211}
\affiliation{%
  \institution{Google LLC}
  \city{Mountain View}
  \state{CA}
  \country{USA}
}
\email{mohitsharma@google.com}

\renewcommand{\shortauthors}{Surendranath Girija et al.}

\begin{abstract}
  Modern production-scale recommender systems rely on complex, multi-task ranking models \cite{10.1145/3580305.3599769, 10.1145/3219819.3220007, 10.1145/3580305.3599881}. Introducing new prediction tasks into these massive systems often causes bottlenecks - it risks negative task conflicts with existing tasks, and can lead to long development and experimentation cycles due to the expensive retraining of backbone models and downstream models or tuning of reward combination formulas. To address the critical challenge of slow experimentation velocity, we introduce the Lightweight Ranking Heads (Light Heads) framework. Designed for continuous online learning environments, Light Heads enable the dynamic injection of new tasks into existing multi-task ranking models, effectively obviating the need for model cold-starting and retraining of backbone models. By utilizing stop-gradients and stateless daily training, this design strictly isolates new tasks, mitigating the risk of adverse task conflicts. Crucially, this framework uses a centralized configuration that allows Light Heads to be added to multiple models simultaneously, unblocking faster training data generation and co-training of downstream models. Successfully deployed at YouTube scale, this approach reduces the iteration cycle for multi-task experimentation from several weeks to days. In this paper, we detail the system architecture, analyze the training dynamics of stateless cold-started heads, compare their performance to full heads, and demonstrate how Light Heads have enabled the rapid A/B experimentation and deployment of new ranking tasks that yield measurable production value.
\end{abstract}




\maketitle

\section{Introduction}
Production-scale recommender systems are usually complex and consist of several dedicated stages for retrieval, ranking, post-ranking, and diversification \cite{10.1145/2959100.2959190}. The ranking stage can consist of large ML models that can optimize for multiple objectives simultaneously \cite{10.1023/A:1007379606734, 10.1145/3580305.3599769, 10.1145/3219819.3220007, 10.1145/3298689.3346997}. The predictions from the ranking stage are combined by hand-tuned reward functions or, more recently, reward models which optimize for desired product objectives \cite{10.1145/3640457.3688184}. This introduces complexity and coupling between the stages of the recommender system, with changes in one stage requiring retraining of or changes in subsequent stages to maintain fidelity. The time taken to run an A/B experiment involving multiple stages increases as a result and slows down the overall experiment velocity. If the ranking models are large backbone models, they also require significant training time and resources to retrain. This makes it challenging to iterate on ranking models, particularly when introducing new ranking tasks. New ranking tasks also have the potential to introduce interactions with existing tasks, which in the best case can lead to positive transfer, which improves the model, and in the worst case can lead to task conflicts that worsen the quality of predictions across tasks \cite{10.5555/3524938.3525784}.

In this paper, we introduce a framework, called Lightweight Ranking Heads or Light Heads, to inject new lightweight heads to multi-task models without needing to retrain the models, and without affecting the quality of the existing tasks. While we present this idea in the context of recommendation/ranking models, this framework can be generalized to other multi-task models. We design the system in a way to use a central configuration for the Light Heads, which is independent of the ranking model code and can be injected into multiple models at the same time in a continual learning setting. This enables serving these Light Heads on more A/B experiments simultaneously, which in turn accelerates experimentation velocity involving downstream models. We discuss the main motivation behind creating this framework, and some of the design considerations that went into it. We also discuss some of the training dynamics and related challenges we observed while implementing this system and run experiments to measure the performance of Light Heads. We conclude by providing the impact of the framework on experimentation velocity and the production impact of deploying heads developed through this framework. Our main contributions are:
\begin{itemize}
\item An extensible framework that uses a central configuration to define new lightweight heads.
\item Dynamic injection of the new heads into the model without cold-starting the entire model.
\item The use of stop-gradients and reset at the start of training runs to maintain training stability and prediction consistency.
\item Sharing the configuration across all models to enable faster experimentation.
\item A robust, production-grade implementation which mitigates the risks of dynamic injection and missing Light Heads.
\end{itemize}
\section{Motivation: Challenges in Developing Production-scale Recommender Systems}
Deploying new prediction tasks in production-scale recommender systems introduces significant bottlenecks that hinder rapid development. This section outlines the critical challenges, ranging from adverse task conflicts to computationally expensive experimentation cycles, that motivated the design of the Lightweight Ranking Heads framework.

\subsection{Considerations When Adding New Ranking Objectives: Task Conflicts and Negative Transfer}

Modern ranking models are generally multi-task systems using shared towers and embedding tables that form the bulk of the model’s parameters \cite{10.1145/3580305.3599769, 10.1145/3580305.3599881}. These objectives are learned by ranking heads which are created on top of this shared architecture. The choice of tasks for these ranking tasks is guided by high-level goals for the system. Similar or aligned tasks can cause the model to learn better shared representations and lead to improved performance on multiple tasks. Conversely, misaligned tasks can create task conflicts which can degrade model quality \cite{10.1145/3383313.3412236, 10.5555/3495724.3496213}. Therefore, adding new ranking tasks to multi-task models requires careful consideration.

One way to avoid task conflicts when adding new ranking objectives is to use stop-gradients with the new tasks, so the shared layers get no gradient updates from the new head during the backward pass. This limits how much the shared towers benefit from the new tasks, but avoids the risk of negative transfer adversely affecting the quality of existing tasks and ensures model stability. In real production systems, tasks may also be added to improve performance on a limited slice of traffic, e.g., based on device, location, age, etc., which may not be beneficial for the majority slice. Stop-gradients may also be beneficial for such tasks.

\subsection{Downstream Dependencies and Experimentation Velocity}

Large-scale recommender systems consist of multiple models operating sequentially before a final recommendation is generated \cite{10.1145/2959100.2959190}. Downstream components, such as reward models or score combination layers, learn directly from the predictions of upstream ranking models being served and require a uniform prediction space for training. Consequently, introducing a new task to an upstream model introduces a new input or a distribution shift for the downstream system. Therefore, evaluating the true, end-to-end impact of a new upstream task requires co-training the downstream model to consume the new predictions and serving the new versions of both the upstream and downstream models.

\begin{figure}[h]
  \centering
  \includegraphics[width=\linewidth]{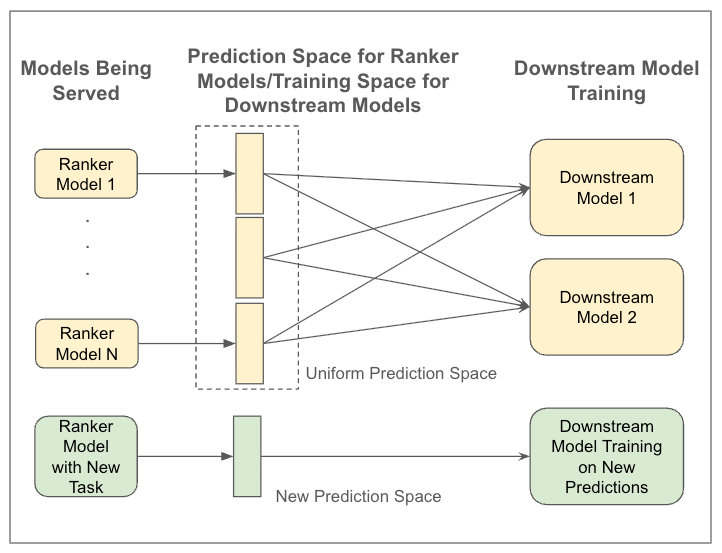}
  \caption{Training downstream models from logged ranking predictions. Adding a new task to a model creates a new prediction space with less data for downstream training}
  \Description{Figure showing ranking models being served which generate new training data for downstream models. One ranking model has a new task and it generates different training data. A custom downstream model is required for learning from this data.}
\end{figure}

The requirement to co-train models slows down experimentation, as each model needs to wait until a sufficient volume of training data is logged from the upstream model’s experiments. Because production platforms run dozens of concurrent A/B experiments, traffic is highly partitioned, and the sample size allocated to any single experiment is limited. Since a model with new tasks generates a new prediction space, and is served in only a few experiments, generating enough training data from this prediction space to effectively co-train a downstream model can take weeks (see Figure \ref{fig:light-heads-experimentation} for experimentation timeline). This tight coupling between sequential stages severely hampers experimentation velocity, as researchers must wait for extended periods to gather sufficient data to train the next model or stage in the system. Our solution provides a way to collect data from several experiments simultaneously, without affecting the fidelity of the experiments. Crucially, by ensuring all upstream models output a uniform prediction space, we unblock downstream models from safely consuming and co-training on this data.

\subsection{Cost of Cold-Starting Backbone Models}
Ranking models can be large backbone models with billions of parameters to learn fine-grained knowledge from varied user behavior and traffic patterns. These models are expensive to train and serve, and require dedicated resources like GPUs or TPUs. When new tasks are added to such models, they are usually cold-started so that the heads and the representations of the shared towers are fully trained for the new tasks. Depending on the model and the amount of training data, this can take days to weeks.

\begin{figure*}[h]
  \centering
  \includegraphics[width=\linewidth]{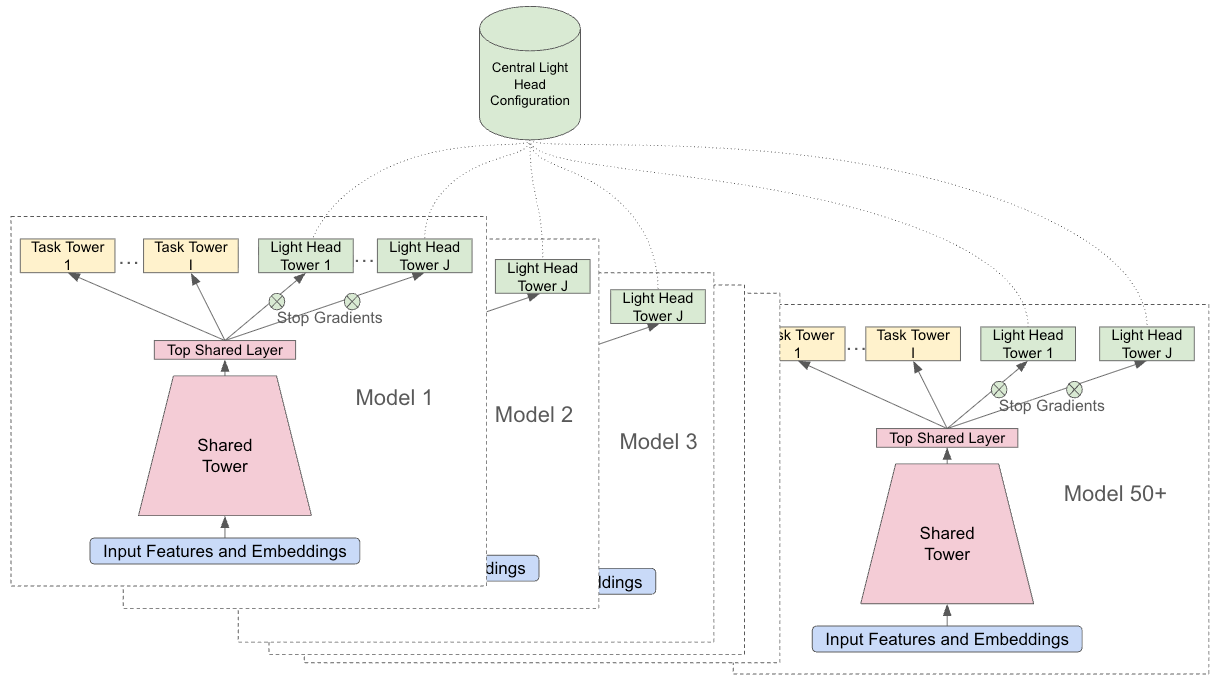}
  \caption{Ranking Model Training Fleet with Light Heads}
  \Description{A recommendation model fleet with each model consisting of a shared tower with multi-task heads and Light Heads at the top. Each model shared Light Heads. The Light Heads are read from a central configuration outside the model and use stop-gradients.}
\end{figure*}

\subsection{Catch-Up Time for Models in Production-scale Systems}
Beyond initial experimentation, deploying new ranking tasks to production introduces additional challenges. Downstream models require a uniform and consistent set of prediction features from upstream models across all traffic to function reliably. However, at any given time, an upstream stage may have hundreds of parallel experimental models running in production. If a new task is added to the primary upstream backbone, the system enters a fragmented state: the newly updated models output predictions for the new task, while the dozens of existing experimental models rely on an older, incompatible set of tasks.

Because downstream systems cannot safely consume fragmented or missing prediction spaces, launching the new tasks becomes blocked. To restore uniformity, every active experimental model in the upstream stage must "catch up", i.e., be updated, branched, and cold-started to incorporate the new tasks. In a production-scale system, waiting for dozens of models to complete this catch-up cycle can delay the deployment of proven improvements by months.

\section{System Architecture: Lightweight Ranking Heads}

\subsection{Core Framework}

The Lightweight Ranking Heads (Light Heads) framework consists of two parts: 
\begin{enumerate}
    \item A central configuration that defines the properties of a ranking head. 
    \item Code within models to consume this configuration file and add Light Heads to their tasks.
\end{enumerate}

\subsubsection{Central Light Head Configuration}
The central configuration for Light Heads consists of a set of configuration files that can be used to define the Light Heads and their details. Engineers/researchers can specify details such as the label, loss function, activation function, regression or classification objectives, metrics for monitoring, and other relevant parameters. They also have the option to extend the existing properties and provide additional functions for sampling or applying custom weights to the data. This configuration becomes the source of truth for all Light Heads used in the system and is stored separately from the code for any single model. It can be used by all ranking models in the system and predictions from the Light Heads can be generated from all models consuming the configuration. This makes it easy for downstream models to be trained faster and speeds up the experimentation velocity significantly. It also eliminates the need for having a large number of models catch-up before the head can be applied in production.

\subsubsection{Integration of Light Heads into Model Architectures}
The Light Heads framework adds a small amount of supporting code to models to read the Light Head configuration, instantiate it and inject it into the model graph at training time. This integration assumes that models support continuous training and can incorporate updates during training iterations. The dynamic injection into existing models solves the problem of having to cold-start the models and saves a significant amount of training resources. While engineers/researchers have flexibility to decide which part of the model the Light Heads are injected into, the standard practice is to inject them at the same layer as existing main heads, so that they benefit from the same shared towers below.

The Light Head architecture comprises a shallow tower with a few hidden layers and is intentionally kept small so that it converges for most tasks within a single training run. In our system, a training run is a scheduled job that processes the latest sequential window of logged data in a continuous online learning setting. There is also some prior literature that supports this choice of smaller heads, such as \cite{10.1145/3485447.3512021}. The framework applies stop-gradients to the bottom of the Light Head layers, which ensures that only the hidden layers of the Light Head are modified during the backward pass, leaving the shared layers below unaffected. This approach is conceptually similar to LoRA \cite{hu2022lora} and Parameter-Efficient Transfer Learning for NLP \cite{houlsby2019parameter}, where adapter layers are added on top of a frozen transformer backbone and sequentially trained for specific tasks. 

The stop-gradients have the following benefits: 
\begin{enumerate}
    \item Light Heads do not cause negative transfer with existing tasks or other Light Heads. 
    \item Gradients introduced by Light Heads, particularly ones that haven’t converged, do not destabilize any layer below them.
\end{enumerate} 

The combination of a shallow architecture and stop-gradients allows Light Heads to be reset at the start of every training run and still converge within the same run. Because the checkpoints for Light Heads are not persisted, they are effectively cold-started with each new iteration. This stateless design ensures that Light Head quality remains predictable and consistent across all models, preventing scenarios where heads in older models become more heavily trained than those in newer models. Consequently, Light Heads exhibit minimal metric variance across the served models, allowing predictions from any upstream model to be safely utilized during serving. This cross-model consistency is an important requirement as it allows the framework to leverage the central configuration to generate uniform predictions from all models simultaneously, thereby significantly increasing the volume of training data for downstream models.

\begin{figure}[h]
  \centering
  \includegraphics[width=\linewidth]{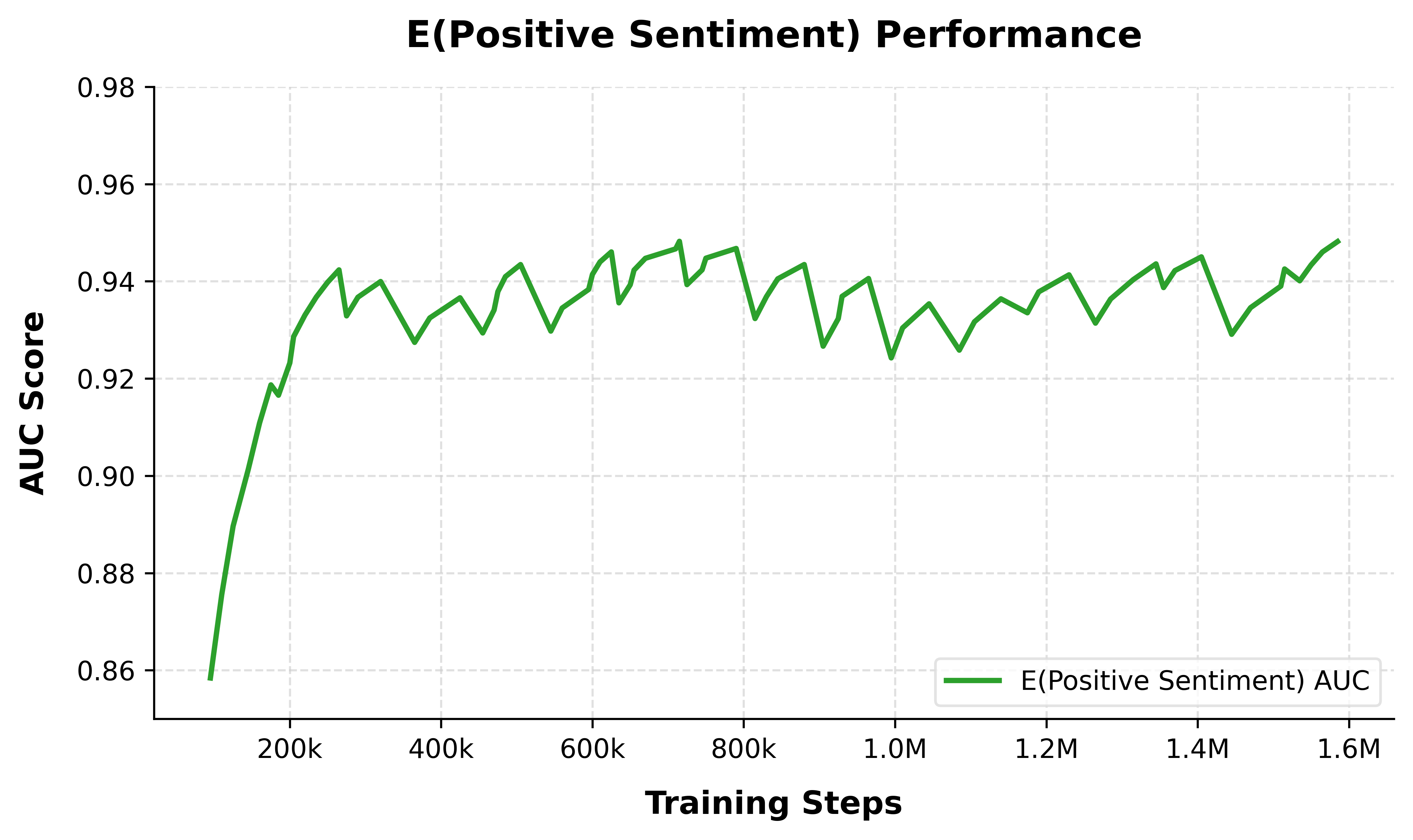}
  \caption{Light Heads Reset at the Start of New Training Runs}
  \Description{An AUC vs step graph tracking the AUC of Light Heads showing a sharp dip in value at the start of a new training run. The Light Head is also able to reach a local optima very quickly with a few steps of training.}
  \label{fig:reset-at-start-of-training}
\end{figure}

The reset at the start of the training run introduces an interesting training dynamic where the metrics of the Light Head drop sharply at the start of the training run and recover quickly within a few steps, as seen in Figure \ref{fig:reset-at-start-of-training}. This also means that the Light Heads rely heavily on the shared tower representation below it, and any improvement in the shared tower directly translates to improvements in the Light Head quality. The quality of Light Head predictions at serving time remains unaffected by the reset as the model is exported for serving only after it completes the entire training run, by which point the Light Heads would have converged to a good local optimum.   

While stop-gradients and the reset at the start of training prevent Light Heads from updating the shared towers, this design leverages the strong generalization capabilities of the pre-existing shared representations. Even though Light Head metrics are slightly lower than their co-trained full head counterparts for dense tasks, our experiments demonstrate that Light Heads can reliably converge to near-parity on sparse tasks. Furthermore, Light Heads allow us to add objectives that would otherwise introduce task conflict without adversely affecting existing tasks or degrading the shared representations. This suggests that for large-scale multi-task models with well-optimized shared towers, the representational capacity is sufficient to support new, isolated tasks without requiring full-model co-training. 

\subsection{Flexibility of the Framework}
The Light Heads framework provides some flexibility to introduce new labels or features that are not consumed by the main model. Researchers/engineers can define new labels in custom data generation jobs, which are joined with the original training data and can have Light Heads which train on these custom labels. This is particularly useful for experimental features or labels that have not been fully verified in production. Users of the framework also have the option to disable resetting specific Light Heads at every new training run, which retains the checkpoints and uses them to warm-start the Light Heads. This has been useful for some very sparse Light Heads where training for longer is beneficial.

The design of the Light Heads framework makes it particularly beneficial for sparse objectives or verticals that the main heads do not specialize well enough on. The main heads in the ranking model train on a large amount of data and sometimes make predictions which can be sub-optimal for very sparse slices of data that have a distinct distribution due to global loss domination. Upsampling or upweighting sparse slices in the main model could have a positive impact on the metrics for the slice, but a negative impact on the overall metrics, so it would not be a worthwhile tradeoff for the main heads. Light Heads can be used to solve this problem as they can be trained only on data from the specific slice without affecting the shared layers. Light Heads have been successfully launched to production in YouTube to improve the recommendations for specific sparse verticals.

\subsection{Production Risks and Mitigations}
The design choice of injecting tasks to all models in a stage of a production recommender system introduces some risks during training and serving, which require additional safeguards. Since models can pick up changes to the central Light Head configuration during a training run, there is a risk of Light Head parameters being modified mid-run, causing the head to become unstable or leading to checkpoint incompatibility. In order to prevent this, the Light Head configuration is read once during the start of the training run and frozen for the duration of that run. Modifications to a Light Head can be picked up during the start of the next run when it will be reset.           

The training data for downstream models consists of Light Head predictions from all models served in production and A/B experiments. Therefore, it is imperative to measure the quality of the Light Heads across the served models, rather than on specific models. We have built a monitoring dashboard to monitor the aggregated quality of the Light Heads and highlight outlier heads before releasing them to production. Furthermore, we run automated evaluations on models before they are exported for serving to ensure that all heads being served, including Light Heads, meet sufficient evaluation thresholds. The framework also allows engineers to define fallback default values for Light Heads which can be used if a Light Head is missing from a model during serving. This makes the system robust to stale models that have not picked up an update to the central Light Head configuration and ensures production stability.     

\section{Experiments}
We evaluate the performance of the Light Heads framework through different experiments comparing Light Heads to full heads in the YouTube Home and Watch Next Ranking models \cite{10.1145/3298689.3346997}. Light Heads are trained with stop-gradients and reset at the start of training runs, while the corresponding full head counterparts are cold-started with the model, and train without stop-gradients for the entire lifetime of the model. We compare the performance of classification and regression Light Heads, and Light Heads with vastly different amounts of training data. We also run ablation studies on stop-gradients and the reset at the start of training runs to show how they affect the quality of Light Heads.

It is important to note that these experiments were conducted at different points in time on a live production system. Because the data is subject to temporal distribution shifts and variations in logged traffic over time, the absolute baseline metrics (e.g., full head AUC) will exhibit slight variations across the different ablation studies. However, the relative comparisons between Light Heads and full heads within each individual table remain controlled.

\subsection{Light Head vs Full Head Performance}\label{sec:offline-comparison}
\subsubsection{Classification Heads}

We compare the performance of various Light Heads with their full head counterparts. The Light Heads are warm-started after 1.2 million training steps, while the full heads start from step 0. When we compare the performance of classification Light Heads in Table \ref{tab:classification-heads}, we see that the Light Heads have AUCs that are slightly lower than their full head counterparts for dense tasks like P(CTR) and E(Interaction Rate). Among the classification heads where the task is sparse, such as E(Positive Sentiment), the Light Head is able to achieve near parity with the full head. This can be attributed to the shared tower before the heads being able to learn good representations which are useful for the task.

\begin{table}[H]
  \caption{Full Head vs Light Head Metrics on Classification Heads}
  \label{tab:classification-heads}
  \begin{tabular}{ccl}
    \toprule
    Head&Full Head AUC&Light Head AUC\\
    \midrule
    P(CTR) & 0.781 & 0.772\\
    E(Positive Sentiment) & 0.9557 & 0.9553\\
    E(Interaction Rate)& 0.9794&0.9704\\
  \bottomrule
\end{tabular}
\end{table}

\subsubsection{Regression Heads}

We also see a similar behavior on regression heads on the RMSE metric. The warm-started Light Head is able to achieve similar metrics as the main head.

\begin{table}[H]
  \caption{Full Head vs Light Head Metrics on Regression Heads}
  \label{tab:regression-heads}
  \begin{tabular}{ccl}
    \toprule
    Head&Full Head RMSE&Light Head RMSE\\
    \midrule
    E(Engagement) & 0.9673 & 0.9675\\
  \bottomrule
\end{tabular}
\end{table}

\subsection{Light Heads without Corresponding Full Heads}
We also conduct experiments to evaluate the performance of standalone Light Heads without a full head equivalent in the same model. This allows us to determine whether the shared representation can generalize well to tasks it is not explicitly trained for. It is important to note that there is some similarity in the overall nature of the task and the training data used for the Light Heads and the full heads. As shown in Table \ref{tab:light-heads-without-full-heads}, Light Heads without corresponding full heads also attain a high value of AUC. 

\begin{table}[H]
  \caption{Light Heads without Corresponding Full Heads}
  \label{tab:light-heads-without-full-heads}
  \begin{tabular}{ccl}
    \toprule
    Head&Light Head AUC\\
    \midrule
    P(Intent to Revisit) & 0.8523\\
    P(Intent to Share) & 0.9224\\
  \bottomrule
\end{tabular}
\end{table}

\begin{figure*}[t]
  \centering
  \includegraphics[width=\textwidth]{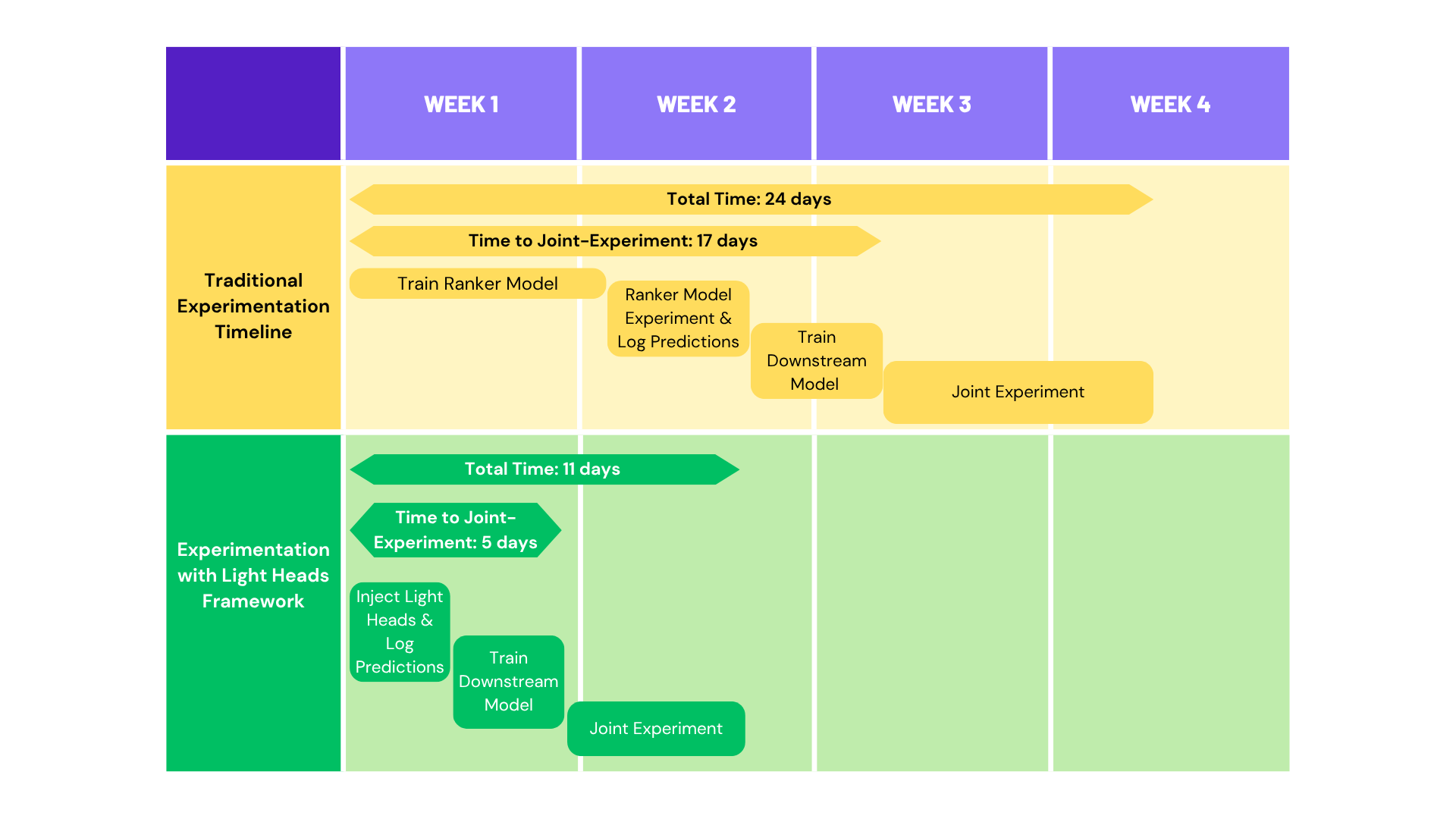}
  \caption{Experimentation Timeline with and without Light Heads}
  \Description{A Gantt chart comparing the experimentation timeline with and without Light Heads.}
  \label{fig:light-heads-experimentation}
\end{figure*}

\subsection{Ablation Study on Stop-Gradients}
To evaluate the efficacy of task isolation, we ablated the stop-gradient operation on the Light Heads and measured the impact on both the full heads and the Light Heads. As shown in Table \ref{tab:ablation-study-full-head-metrics}, removing the stop-gradients degraded the metrics of the full heads. This effect is more pronounced on heads that have more training data.

\begin{table}[H]
  \caption{Full Head Metrics with Light Head Stop-Gradient Ablation}
  \label{tab:ablation-study-full-head-metrics}
  \small
  \begin{tabular}{ccl}
    \toprule
    Full Head&With Stop Grads&Ablate Stop Grads\\
    \midrule
    P(CTR) AUC& 0.7767 & 0.7704\\
    E(Positive Sentiment) AUC& 0.9557 & 0.9541\\
    E(Engagement) RMSE& 0.9887 & 0.9975\\
  \bottomrule
\end{tabular}
\end{table}

We also evaluate the metrics of Light Heads when we ablate the stop-gradients. Surprisingly, some Light Heads have worse quality without stop-gradients, as their corresponding full head counterparts in the same model also get worse. This can be attributed to the shared tower learning a slightly worse representation than when the Light Heads had stop-gradients. This supports our premise that adding new tasks can lead to conflict between tasks if they are not aligned.

\begin{table}[H]
  \caption{Light Head Metrics with Light Head Stop Gradient Ablation}
  \label{tab:ablation-study-light-head-metrics}
  \small
  \begin{tabular}{ccl}
    \toprule
    Light Head&With Stop Grads&Ablate Stop Grads\\
    \midrule
    P(CTR) AUC& 0.7667 & 0.7584\\
    E(Positive Sentiment) AUC& 0.9553 & 0.9541\\
    E(Engagement) RMSE& 0.9886 & 0.9976\\
  \bottomrule
\end{tabular}
\end{table}

\subsection{Ablation Study on Reset at the Start of Training Run}
When we ablate the reset at the start of training, Light Heads  are able to attain a slightly better optimum than with the reset. This is particularly beneficial for Light Heads that have a lot less training data. So, engineers/researchers have the option of disabling reset when they define the  Light Heads.

\begin{table}[H]
  \caption{Light Head Metrics with Light Head Reset Ablation}
  \label{tab:regression_heads}
  \small
  \begin{tabular}{ccl}
    \toprule
    Light Head&With Reset&Without Reset\\
    \midrule
    P(CTR) AUC& 0.764 & 0.7646\\
    E(Positive Sentiment) AUC& 0.9508 & 0.952\\
    E(Engagement) RMSE& 0.9884 & 0.9864\\
  \bottomrule
\end{tabular}
\end{table}

\section{Production Impact}
The Light Heads framework has been applied successfully within YouTube on two large model families and has significantly reduced the experimentation duration when adding new heads. The total time for an experiment that involves the ranker model and a downstream model has been reduced from 24 days to 11 days. Crucially, the time for starting a joint experiment has been reduced from 17 days to 5 days. In the context of a large platform running dozens of parallel experiments, this leads to substantial savings in development time and resources. Furthermore, Light Heads launched to production have also brought about improvements in the top-line metrics and vertical-specific metrics.

\subsection{Effect on Experiment Velocity}
The experimentation cycle involving a ranking model and a downstream model, such as a learned-ranking model, is particularly time consuming as it involves several steps and takes multiple weeks, as shown in the top half of Figure \ref{fig:light-heads-experimentation}. The training of the main ranker model can take one week to several weeks, after which it is used in an A/B experiment to generate predictions.

These predictions will be logged and used to train downstream models. However, sufficient data needs to be generated from the ranker A/B experiment before training the downstream model and can take a few days. Only once the downstream model is fully trained can a joint experiment involving both the ranker model and the downstream model be started. In total, this experiment cycle can span four weeks or more.

When using the Light Heads framework, the Light Heads are injected into continuously trained ranker models and are ready for serving in as little as one day. The Light Heads are also picked up by all ranker models through the central configuration. Training data containing predictions from these Light Heads are generated by all models in production and A/B experiments, vastly increasing the volume of training data, thereby enabling downstream models to be trained faster. The joint experiment with the downstream model and the ranker with the Light Head can start in as little as five days, bringing down the overall experimentation time significantly. Figure \ref{fig:light-heads-experimentation} illustrates the two timelines.

The timeline for launching an approved full head to production is also lengthy as models in experiments need to serve the new head to generate a uniform prediction distribution for downstream models. This can take two to three months, accounting for the time required for all experimental models to adopt the new head. Using Light Heads significantly shortens this time as the Light Heads are warm-started and don’t need the experimental models to be restarted. This can bring down the launch time to a few days, after the quality benefits of a Light Head have been validated in A/B experiments. Because the engineering effort and compute savings are substantial, successful Light Heads are typically retained in production rather than graduating them into full heads. We have found that the massive reduction in deployment time and system complexity outweighs the slight offline metric degradation observed on dense tasks.

\subsection{Effect on Metrics}
The Light Head framework has been successfully used to launch new objectives to production, yielding positive impacts on key YouTube metrics. Two illustrative examples highlight this success. First, a Light Head for optimizing long-horizon rewards drove a statistically significant  ($p<0.05$) +0.03\% improvement in top-line engagement and a 0.40\% reduction in low-quality impressions. The reduction in low-quality impressions ensures users are recommended higher-quality content, directly contributing to long-term user satisfaction. Second, a Light Head tailored for users with a paid subscription to Primetime channels for TV shows, movies and live events \cite{primetimechannels}, which resulted in a significant +13.83\% improvement in primetime-specific engagement. Furthermore, by utilizing the Light Heads framework, the end-to-end development of these impactful objectives was achieved while saving months of engineering effort and substantial TPU-time.

\section{Limitations}
While the Light Heads framework significantly accelerates experimentation velocity and reduces training compute required at YouTube-scale, its generalizability to other production environments depends on a few architectural and infrastructure prerequisites.

\subsubsection*{Infrastructure and Scale Prerequisites:}
The framework's design makes it suitable for continual learning systems. It relies heavily on a continuous training pipeline where models can dynamically load centralized configurations and frequently ingest new data. For organizations and systems that rely on infrequent batch training, the stateless reset mechanism would not be viable, as the Light Heads require sufficient continuous training steps to converge before serving. Furthermore, the engineering overhead of building and maintaining the Light Heads framework is best justified in environments with a large number of models training in parallel and running dozens of concurrent A/B experiments. In smaller-scale systems where the prediction space fragmentation problem is minimal, a traditional cold-start approach is more reasonable.

\subsubsection*{Architectural Dependencies:}
The experimentation time savings of the framework are fully realized in multi-stage recommender systems where downstream models rely on upstream model predictions. In single-stage recommender systems, such as \cite{deng2025onerec}, there is no downstream data dependency. In such systems, the benefit of Light Heads would be primarily limited to the compute savings from not having to retrain the backbone model.

\subsubsection*{Cost vs Quality Trade-off:}
Finally, as noted in offline evaluation in Section \ref{sec:offline-comparison}, Light Heads exhibit a slight offline metric degradation on dense tasks compared to fully co-trained heads. The framework operates on a cost-vs-quality Pareto trade-off, i.e., it is acceptable to have slight degradation in prediction accuracy in exchange for a significant reduction in experimentation time and compute savings. Systems that are sensitive to even minor degradation in quality on dense tasks, and where compute costs and time to experimentation are not limiting factors, may be better suited for cold-starting models with the new tasks added as full heads. 

\section{Conclusion and Future Work}
In this paper, we have provided an overview of the Lightweight Ranking Heads (Light Heads) framework in YouTube recommendation systems. We outline the motivation behind the framework and provide details about the core framework components. We go over specific design choices like stop-gradients and reset at the start of training runs in a continual online learning setting, discuss the tradeoffs involved, and mention the steps taken to make the framework production-ready. The experiments in the paper compare the performance of Light Heads to full heads across tasks, and measure the effect of ablating key choices made by the framework. Finally, we show how the framework significantly accelerated experiment velocity within YouTube and led to impactful production improvements.

\printbibliography

@inproceedings{10.1145/3219819.3220007,
author = {Ma, Jiaqi and Zhao, Zhe and Yi, Xinyang and Chen, Jilin and Hong, Lichan and Chi, Ed H.},
title = {Modeling Task Relationships in Multi-task Learning with Multi-gate Mixture-of-Experts},
year = {2018},
isbn = {9781450355520},
publisher = {Association for Computing Machinery},
address = {New York, NY, USA},
url = {https://doi.org/10.1145/3219819.3220007},
doi = {10.1145/3219819.3220007},
booktitle = {Proceedings of the 24th ACM SIGKDD International Conference on Knowledge Discovery \& Data Mining},
pages = {1930–1939},
numpages = {10},
location = {London, United Kingdom},
series = {KDD '18}
}

@inproceedings{10.1145/2959100.2959190,
author = {Covington, Paul and Adams, Jay and Sargin, Emre},
title = {Deep Neural Networks for YouTube Recommendations},
year = {2016},
isbn = {9781450340359},
publisher = {Association for Computing Machinery},
address = {New York, NY, USA},
url = {https://doi.org/10.1145/2959100.2959190},
doi = {10.1145/2959100.2959190},
booktitle = {Proceedings of the 10th ACM Conference on Recommender Systems},
pages = {191–198},
numpages = {8},
location = {Boston, Massachusetts, USA},
series = {RecSys '16}
}

@inproceedings{10.1145/3298689.3346997,
author = {Zhao, Zhe and Hong, Lichan and Wei, Li and Chen, Jilin and Nath, Aniruddh and Andrews, Shawn and Kumthekar, Aditee and Sathiamoorthy, Maheswaran and Yi, Xinyang and Chi, Ed},
title = {Recommending what video to watch next: a multitask ranking system},
year = {2019},
isbn = {9781450362436},
publisher = {Association for Computing Machinery},
address = {New York, NY, USA},
url = {https://doi.org/10.1145/3298689.3346997},
doi = {10.1145/3298689.3346997},
booktitle = {Proceedings of the 13th ACM Conference on Recommender Systems},
pages = {43–51},
numpages = {9},
location = {Copenhagen, Denmark},
series = {RecSys '19}
}

@inproceedings{10.1145/3383313.3412236,
author = {Tang, Hongyan and Liu, Junning and Zhao, Ming and Gong, Xudong},
title = {Progressive Layered Extraction (PLE): A Novel Multi-Task Learning (MTL) Model for Personalized Recommendations},
year = {2020},
isbn = {9781450375832},
publisher = {Association for Computing Machinery},
address = {New York, NY, USA},
url = {https://doi.org/10.1145/3383313.3412236},
doi = {10.1145/3383313.3412236},
booktitle = {Proceedings of the 14th ACM Conference on Recommender Systems},
pages = {269–278},
numpages = {10},
location = {Virtual Event, Brazil},
series = {RecSys '20}
}

@inproceedings{10.5555/3495724.3496213,
author = {Yu, Tianhe and Kumar, Saurabh and Gupta, Abhishek and Levine, Sergey and Hausman, Karol and Finn, Chelsea},
title = {Gradient surgery for multi-task learning},
year = {2020},
isbn = {9781713829546},
publisher = {Curran Associates Inc.},
address = {Red Hook, NY, USA},
booktitle = {Proceedings of the 34th International Conference on Neural Information Processing Systems},
articleno = {489},
numpages = {13},
location = {Vancouver, BC, Canada},
series = {NIPS '20}
}

@inproceedings{10.1145/3640457.3688184,
author = {Wu, Yi and Chang, Daryl and She, Jennifer and Zhao, Zhe and Wei, Li and Heldt, Lukasz},
title = {Learned Ranking Function: From Short-term Behavior Predictions to Long-term User Satisfaction},
year = {2024},
isbn = {9798400705052},
publisher = {Association for Computing Machinery},
address = {New York, NY, USA},
url = {https://doi.org/10.1145/3640457.3688184},
doi = {10.1145/3640457.3688184},
booktitle = {Proceedings of the 18th ACM Conference on Recommender Systems},
pages = {1004–1009},
numpages = {6},
location = {Bari, Italy},
series = {RecSys '24}
}

@inproceedings{10.1145/3485447.3512021,
author = {Wang, Yuyan and Zhao, Zhe and Dai, Bo and Fifty, Christopher and Lin, Dong and Hong, Lichan and Wei, Li and Chi, Ed H.},
title = {Can Small Heads Help? Understanding and Improving Multi-Task Generalization},
year = {2022},
isbn = {9781450390965},
publisher = {Association for Computing Machinery},
address = {New York, NY, USA},
url = {https://doi.org/10.1145/3485447.3512021},
doi = {10.1145/3485447.3512021},
booktitle = {Proceedings of the ACM Web Conference 2022},
pages = {3009–3019},
numpages = {11},
location = {Virtual Event, Lyon, France},
series = {WWW '22}
}

@inproceedings{10.1145/3580305.3599769,
author = {Li, Danwei and Zhang, Zhengyu and Yuan, Siyang and Gao, Mingze and Zhang, Weilin and Yang, Chaofei and Liu, Xi and Yang, Jiyan},
title = {AdaTT: Adaptive Task-to-Task Fusion Network for Multitask Learning in Recommendations},
year = {2023},
isbn = {9798400701030},
publisher = {Association for Computing Machinery},
address = {New York, NY, USA},
url = {https://doi.org/10.1145/3580305.3599769},
doi = {10.1145/3580305.3599769},
booktitle = {Proceedings of the 29th ACM SIGKDD Conference on Knowledge Discovery and Data Mining},
pages = {4370–4379},
numpages = {10},
location = {Long Beach, CA, USA},
series = {KDD '23}
}

@inproceedings{10.1145/3580305.3599881,
author = {Tan, Chun How and Chan, Austin and Haldar, Malay and Tang, Jie and Liu, Xin and Abdool, Mustafa and Gao, Huiji and He, Liwei and Katariya, Sanjeev},
title = {Optimizing Airbnb Search Journey with Multi-task Learning},
year = {2023},
isbn = {9798400701030},
publisher = {Association for Computing Machinery},
address = {New York, NY, USA},
url = {https://doi.org/10.1145/3580305.3599881},
doi = {10.1145/3580305.3599881},
booktitle = {Proceedings of the 29th ACM SIGKDD Conference on Knowledge Discovery and Data Mining},
pages = {4872–4881},
numpages = {10},
location = {Long Beach, CA, USA},
series = {KDD '23}
}

@misc{primetimechannels,
  author = {Erin Teague},
  title = {Get more of your favorite content on YouTube with Primetime Channels},
  year = {2022},
  month = {November},
  howpublished = {YouTube Official Blog},
  url = {https://blog.youtube/news-and-events/more-of-your-favorite-content-on-youtube-with-primetime-channels},
  note = {Accessed: May 19, 2026}
}

@article{10.1023/A:1007379606734,
author = {Caruana, Rich},
title = {Multitask Learning},
year = {1997},
issue_date = {July 1997},
publisher = {Kluwer Academic Publishers},
address = {USA},
volume = {28},
number = {1},
issn = {0885-6125},
url = {https://doi.org/10.1023/A:1007379606734},
doi = {10.1023/A:1007379606734},
journal = {Mach. Learn.},
month = jul,
pages = {41–75},
numpages = {35}
}

@inproceedings{10.5555/3524938.3525784,
author = {Standley, Trevor and Zamir, Amir and Chen, Dawn and Guibas, Leonidas and Malik, Jitendra and Savarese, Silvio},
title = {Which tasks should be learned together in multi-task learning?},
year = {2020},
publisher = {JMLR.org},
booktitle = {Proceedings of the 37th International Conference on Machine Learning},
articleno = {846},
numpages = {13},
series = {ICML'20}
}

@article{hu2022lora,
  title={Lora: Low-rank adaptation of large language models.},
  author={Hu, Edward J and Shen, Yelong and Wallis, Phillip and Allen-Zhu, Zeyuan and Li, Yuanzhi and Wang, Shean and Wang, Liang and Chen, Weizhu and others},
  journal={Iclr},
  volume={1},
  number={2},
  pages={3},
  year={2022}
}

@inproceedings{houlsby2019parameter,
  title={Parameter-efficient transfer learning for NLP},
  author={Houlsby, Neil and Giurgiu, Andrei and Jastrzebski, Stanislaw and Morrone, Bruna and De Laroussilhe, Quentin and Gesmundo, Andrea and Attariyan, Mona and Gelly, Sylvain},
  booktitle={International conference on machine learning},
  pages={2790--2799},
  year={2019},
  organization={PMLR}
}

@article{deng2025onerec,
  title={Onerec: Unifying retrieve and rank with generative recommender and iterative preference alignment},
  author={Deng, Jiaxin and Wang, Shiyao and Cai, Kuo and Ren, Lejian and Hu, Qigen and Ding, Weifeng and Luo, Qiang and Zhou, Guorui},
  journal={arXiv preprint arXiv:2502.18965},
  year={2025}
}

\end{document}